\documentclass[sigconf]{acmart}
\AtBeginDocument{%
  }

\setcopyright{acmlicensed}
\copyrightyear{2025}
\acmYear{2025}
\acmDOI{10.1145/3696410.3714518}
\setcopyright{acmlicensed}\acmConference[WWW '25]{Proceedings of the ACM
Web Conference 2025}{April 28-May 2, 2025}{Sydney, NSW, Australia}
\acmBooktitle{Proceedings of the ACM Web Conference 2025 (WWW '25), April
28-May 2, 2025, Sydney, NSW, Australia}

\acmISBN{979-8-4007-1274-6/25/04}

\begin{document}

\title{Generative Dynamic Graph Representation Learning for Conspiracy Spoofing Detection}


\author{Sheng Xiang}
\orcid{0000-0002-1454-8240}
\affiliation{%
 \institution{Tongji University}
 \city{Shanghai}
 \country{China}
}
\email{xiangsheng218@gmail.com}

\author{Yidong Jiang}
\orcid{0009-0000-4033-8709}
\affiliation{%
 \institution{Tongji University}
 \city{Shanghai}
 \country{China}
}
\email{2253899@tongji.edu.cn}

\author{Yunting Chen}
\orcid{0009-0006-5228-0464}
\affiliation{%
 \institution{University of Technology Sydney}
 \city{Sydney}
 \state{NSW}
 \country{Australia}
}
\email{yunting.chen@uts.edu.au}

\author{Dawei Cheng}
\orcid{0000-0002-5877-7387}
\authornote{Corresponding author: Dawei Cheng.}
\affiliation{%
  \institution{Tongji University \& Shanghai Artificial Intelligence Laboratory}
  \city{Shanghai}
  \country{China}
  }
\email{dcheng@tongji.edu.cn}

\author{Guoping Zhao}
\orcid{0009-0000-3032-9825}
\affiliation{%
\institution{China Futures Market Monitoring Center}
\city{Beijing}
\country{China}}
\email{zhaogp@cfmmc.com}

\author{Changjun Jiang}
\orcid{0000-0003-0637-9317}
\affiliation{%
  \institution{Tongji University \& Shanghai Artificial Intelligence Laboratory}
  \city{Shanghai}
  \country{China}
  }
\email{cjjiang@tongji.edu.cn}

\renewcommand{\shortauthors}{Sheng Xiang et al.}

\begin{abstract}
Spoofing detection in financial trading is crucial, especially for identifying complex behaviors such as conspiracy spoofing. Traditional machine-learning approaches primarily focus on isolated node features, often overlooking the broader context of interconnected nodes. Graph-based techniques, particularly Graph Neural Networks (GNNs), have advanced the field by leveraging relational information effectively. However, in real-world spoofing detection datasets, trading behaviors exhibit dynamic, irregular patterns. Existing spoofing detection methods, though effective in some scenarios, struggle to capture the complexity of dynamic and diverse, evolving inter-node relationships. To address these challenges, we propose a novel framework called the Generative Dynamic Graph Model (GDGM), which models dynamic trading behaviors and the relationships among nodes to learn representations for conspiracy spoofing detection. Specifically, our approach incorporates the generative dynamic latent space to capture the temporal patterns and evolving market conditions. Raw trading data is first converted into time-stamped sequences. Then we model trading behaviors using the neural ordinary differential equations and gated recurrent units, to generate the representation incorporating temporal dynamics of spoofing patterns. Furthermore, pseudo-label generation and heterogeneous aggregation techniques are employed to gather relevant information and enhance the detection performance for conspiratorial spoofing behaviors. Experiments conducted on spoofing detection datasets demonstrate that our approach outperforms state-of-the-art models in detection accuracy. Additionally, our spoofing detection system has been successfully deployed in one of the largest global trading markets, further validating the practical applicability and performance of the proposed method.
\end{abstract}



\maketitle

\section{Introduction}
\label{sec:intro}
\begin{figure}[t]
  \centering
  \includegraphics[width=\linewidth]{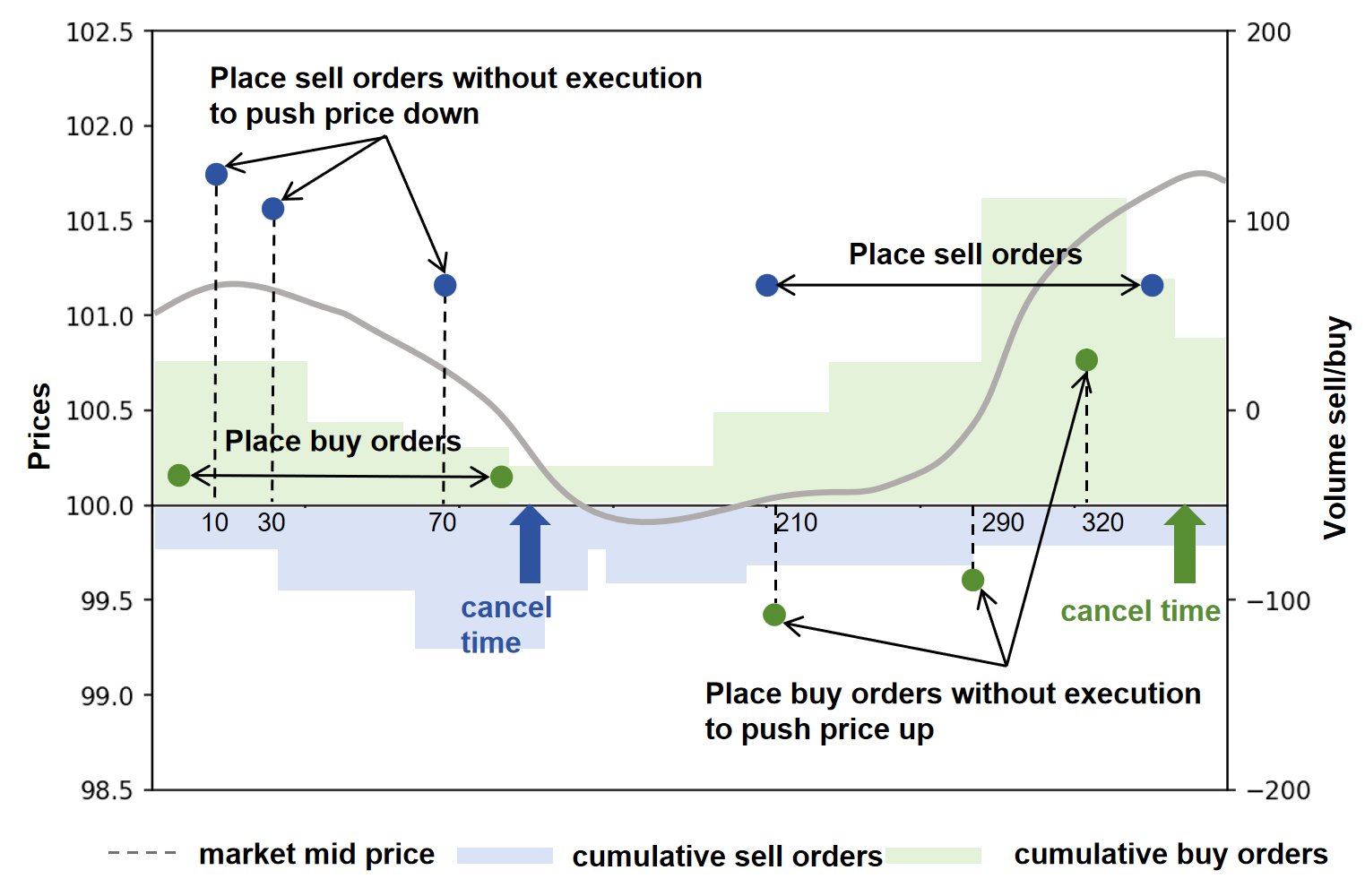}
  \caption{A typical example of spoofing transactions. Traders place deceptive sell (ask) or buy (bid) orders without execution to influence the market price by misleading other traders about the market demand or supply.}
  \Description{A typical example of spoofing transactions.}\label{fig:spoof}
\end{figure}

Spoofing transaction \cite{7445815}, commonly known as ``deceptive trading'' in financial markets, represents a sophisticated form of market manipulation characterized by strategic order placement. As illustrated in Figure~\ref{fig:spoof}, traders intentionally submit non-executable orders (e.g., large-volume bids or asks in a small period) to fabricate artificial market signals about supply/demand equilibrium. This manipulative practice distorts asset pricing mechanisms for stocks, currencies, and derivatives by creating illusory liquidity or volatility patterns that enable illicit gains through subsequent counter-trades \cite{7454730}. Functioning as a deceptive market intervention, spoofing behavior erodes market fairness and efficiency, contributing to systemic risks and substantial investor losses \cite{Bolton2002StatisticalFD,cheng2022financial}. The proliferation of algorithmic trading platforms has exponentially amplified both the occurrence rate and economic consequences of spoofing activities, compelling global regulators to implement stringent surveillance frameworks and punitive measures \cite{9243545,cheng2023anti}. Consequently, developing advanced detection paradigms for spoofing transactions has emerged as a paramount research priority across financial trading institutions and academic communities.

Early systems for detecting spoofing in financial transactions relied heavily on rule-based methods, where alerts were triggered by predefined behavioral thresholds \cite{Whitrow2009TransactionAA,xiang2022efficient,xiang2022general,xiang2025scalable}. While effective in static scenarios, these methods lacked the adaptability needed for the evolving tactics of fraudsters. Machine learning and deep learning approaches have since emerged as more dynamic solutions, offering data-driven methodologies that continuously refine detection capabilities \cite{bhattacharyya2011data}. Techniques such as convolutional neural networks (CNNs)~\cite{fu2016credit}, recurrent neural networks (RNNs)~\cite{xiang2022temporal,jiang2025generative}, and attention mechanisms~\cite{cheng2020spatio,Xiang2023SemiSupervisedCC} have been utilized to capture complex transaction patterns, marking significant progress in fraud detection \cite{doi:10.1080/1350486X.2020.1726783}. However, these methods often fall short of capturing the intricate relationships and dynamics within interconnected transactions, ignoring the rich information from relations.

Graph-based methods, particularly Graph Neural Networks, have transformed fraud detection by effectively leveraging relational structures and connectivity patterns among transactions \cite{cheng2020graph}. GNNs, such as CARE-GNN and SemiGNN, have demonstrated notable success in modeling dependencies across nodes and incorporating both relational and temporal information for improved detection performance \cite{10.1145/3340531.3411903,Wang2019ASG}. Recent advancements have introduced models like RTG-Trans, which integrates deep graph learning with temporal analysis to enhance spoofing detection accuracy \cite{RTG-Trans}. Additionally, GPEGNN combines local and global community features to better identify and analyze conspiracy spoofing behaviors \cite{GPEGNN}.

Despite the significant progress made by machine learning and graph-based approaches in detecting spoofing trading, several limitations remain that hinder their applicability to real-world financial trading data. First, the temporal relationships between transactions are highly irregular, with varying distances and dependencies that cannot be effectively modeled by trivial recurrent neural networks (RNNs) or standard transformer architectures. These methods rely on assumptions of regular or sequential data patterns, e.g., words and image patches, making them ill-suited for capturing the dynamic and erratic nature of trading behaviors. Second, the structural relationships in transaction networks exhibit inherent heterogeneity, as the interactions between transactions form non-homophily graph structures~\cite{zou2024effective}. Conventional Graph Neural Networks (GNNs), such as Graph Convolutional Networks (GCNs) and Graph Attention Networks (GATs), are limited in their ability to distinguish between non-homophily graph representations, leading to reduced performance when analyzing the diverse and complex patterns required for conspiracy spoofing detection.

To address these challenges, we proposed Generative Dynamic Graph Model (GDGM). Specifically, our method comprises four steps: First, we convert raw trading data into structured time-stamped sequences for generative dynamic data encoding. During encoding, we introduce neural ordinary differential equation (Neural ODE) as the latent space generative method to obtain temporal representations from transaction sequences. Second, we introduce a Transaction Graph Pseudo-Label Generation mechanism to assign pseudo-labels for unlabeled nodes, thereby improving the model's capability to better identify patterns associated with conspiracy spoofing transactions. Finally, heterogeneous aggregation integrates different types of information across the graph, enabling the model to capture the diverse and non-homophily graph structures inherent in trading data. The experiments conducted on real-world spoofing detection datasets demonstrate that GDGM achieves superior detection performance when compared to existing state-of-the-art methods. Furthermore, the deployment of our system in one of the largest global trading markets, coupled with a case study, highlights the practical effectiveness of GDGM in addressing conspiracy spoofing detection.

The contributions of our work can be summarized as follows:

\begin{itemize}
    \item We transform raw trading data into time-stamped sequences. Then, by utilizing generative neural ordinary differential equations, our method effectively captures temporal dependencies in transaction representations.

    \item To handle the non-homophily and diverse relationships in trading data, we generate pseudo-labels and design a heterogeneous aggregation mechanism. This mechanism enables our model to adaptively integrate relational information across different types of connections in trading data.

    \item Experimental results on real-world spoofing detection datasets demonstrate that GDGM outperforms state-of-the-art models in terms of detection accuracy. Furthermore, our method has been successfully deployed in one of the largest global trading markets, and a case study highlights its superior capability in uncovering spoofing patterns.
\end{itemize}

\section{Related Works}
This section reviews existing methods for spoofing detection and graph learning on financial transactions, highlighting their key contributions and main limitations.

\subsection{Spoofing Detection}
Since spoofing poses a major threat to the stability of financial markets, a variety of detection methodologies have been introduced in recent years to detect this issue. Early approaches mainly relied on statistical analysis of transaction behaviors to identify spoofing patterns \cite{6803980}. Machine learning-based methods, such as Linear Regression~\cite{LR}, Decision Tree~\cite{RF,GBDT} and Multi-Layer Perceptron~\cite{HMLP} were broadly implemented in real-world spoofing detection systems. As spoofing tactics evolved, researchers started to adopt deep learning techniques, such as recurrent neural networks (RNNs) and attention-based neural networks, to capture more complex trading patterns\cite{doi:10.1080/1350486X.2020.1726783,Arora2019FingerprintSD}. While these approaches offered advancements, they still had limitations in capturing the intricate interactions within transactions.
More recently, graph neural networks (GNNs) have gained attention for their effectiveness in spoofing detection by leveraging inter-connected behavior within transaction graphs \cite{10.1145/3340531.3411903}. For instance, RTG-Trans combines deep graph learning with temporal analysis to enhance spoofing detection \cite{RTG-Trans}. However, many GNN models focus primarily on local context, relying on adjacent nodes, which restricts their ability to identify complex, coordinated spoofing activities that require a global perspective on transaction relationships.
Apart from graph-based techniques, other approaches have also been developed to tackle spoofing detection from different angles. Rule-based algorithms, for instance, set specific parameters to identify suspicious trading patterns. A good example of this can be found in studies analyzing transactions across stocks in indices like the Ibovespa, as shown in \cite{Mendona2020DetectionAA}. Moreover, researchers have also explored spoofing from a micro-structural perspective. They have introduced variables such as multilevel imbalances in price action and delved into the optimization strategies that potential spoofers might employ, as described in \cite{doi:10.1080/14697688.2022.2059390}. Despite these continuous advancements, existing methods still struggle when it comes to capturing the temporal relationships in trading behaviors. These relationships are inherently dynamic and do not fit well with traditional RNNs or transformer-based approaches.

\subsection{Graph Learning on Financial Transaction}
Graph-based machine learning has become instrumental in analyzing financial transactions and detecting fraudulent activities, drawing on its effectiveness in fields such as image processing, natural language processing, and knowledge graph construction\cite{10.5555/3294771.3294869,10.1145/3502289,cheng2025graph,zou2024subgraph,cheng2022regulating}. Within financial applications, graph models have proven adept at addressing complex tasks like credit risk assessment~\cite{liu2023transformed} and financial fraud detection~\cite{li2024relation}. For example, vulnerable nodes on the guarantee loan network can be detected by graph models~\cite{cheng2021efficient}. One significant challenge in credit risk assessment for small and medium-sized enterprises is the limited sample size. To address this, Wang et al. proposed an adaptive heterogeneous multi-view graph learning model, integrating multiple data perspectives to aggregate heterogeneous information for a comprehensive evaluation of credit risks\cite{10.1155/2021/6670873}. Based on the idea of leveraging diverse data sources, SemiGNN utilizes both labeled and unlabeled data to capture dependencies across data views and neighboring nodes through a hierarchical attention mechanism \cite{Wang2019ASG}. Another major challenge is that fraudulent activities often involve sophisticated tactics such as disguising features and relationships within the data. To address this, Dou et al. introduced CARE-GNN, which incorporates a label-aware similarity measure and a reinforcement-learning-driven neighbor selection strategy, dynamically focusing on relevant nodes based on label information to improve detection accuracy\cite{10.1145/3340531.3411903}. In addition to relational data, temporal patterns play a crucial role in identifying anomalies~\cite{ma2024parallel}. GADBench offers a benchmark framework for sequence-based anomaly detection, capturing user behavior patterns over time, facilitating the detection of anomalies within temporal transaction data\cite{tang2023gadbench}. While these approaches address challenges in fraud detection broadly, they neglect the specific requirements of spoofing detection, particularly the need to distinguish between non-homophily graph structures. To date, no dedicated solutions have been proposed to model the heterogeneous graph relationships unique to spoofing scenarios.

\begin{figure*}[t]
    \centering
    \includegraphics[width=\linewidth]{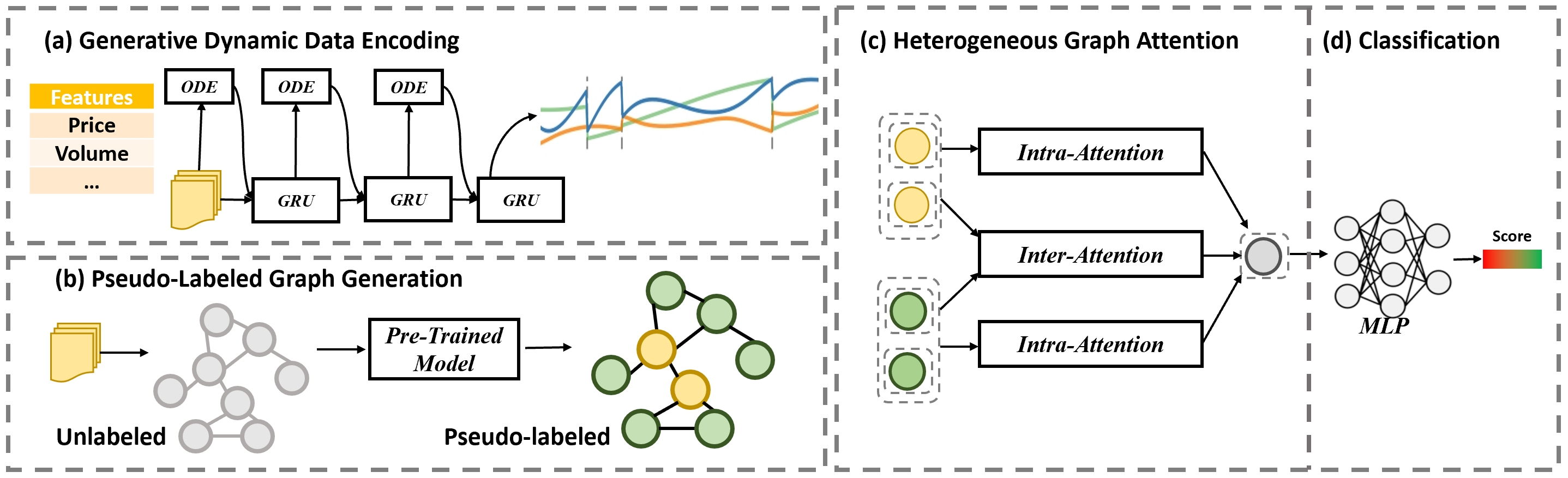}
    \caption{The proposed Generative Dynamic Graph Model (GDGM) architecture for Conspiracy Spoofing Detection. The first part is the historical transaction encoding of input time series data, which builds the embedding of irregular transaction series. The second part is the temporal graph attention. The third part is the heterogeneous graph attention layer, which aggregates the information for different types of neighbors. The fourth part is the classification layer, which is a multi-layer perception that gives the prediction of whether this transaction is spoofing.}
    \label{fig:frame}
\end{figure*}

\section{Methodology}
In this section, according to Figure~\ref{fig:frame}, the proposed framework, GDGM, consists of four main components: (1) Generative Dynamic Data Encoding, (2) Pseudo-Labeled Graph Generation, (3) Heterogeneous Graph Attention, and (4) Classification. These components work together to capture generative dynamics data representations and heterogeneous graph relationships in financial transaction data, effectively modeling conspiracy spoofing behaviors.

\begin{table}[t]
\centering
\caption{Notations used in this paper.}
\begin{tabular}{|c|p{5cm}|}
\hline
\textbf{Symbol} & \textbf{Definition} \\ \hline
$n$ & the total number of nodes \\ \hline
$m$ & the total number of edges \\ \hline
$r$ & the number of relationships in graph $\tilde{G}$ \\ \hline
$d$ & the number of dimension \\ \hline
$Z_i$ & the transformed representation using the $i$-th wavelet kernel \\ \hline
$H$ & the comprehensive node representation aggregated from $Z_0$ to $Z_C$ \\ \hline
$p_i$ & the anomaly probability for node $i$ \\ \hline
$X = \{x_1, x_2, ..., x_n\}$ & the set of node features \\ \hline
$\hat{y}_i$ & the pseudo-label for node $i$ \\ \hline
\end{tabular}
\end{table}

\subsection{Generative Dynamic Data Encoding}

To effectively model the irregular temporal dynamics inherent in financial transaction data, especially in spoofing detection, we first explicitly leverage the timestamp of each transaction. Then, we employ a generative representation learning method, namely Neural Ordinary Differential Equation-based Recurrent Neural Network (ODERNN). This approach combines the continuous-time generative modeling capability of ODEs with the sequential representation power of GRU cells. The encoding process captures both temporal irregularities and sequential dependencies in the trading data.

\subsubsection{Neural ODE Dynamics}
We set the initial hidden state $h(0)$ as zero tensors. Then, the temporal evolution of the hidden state \( h(t) \) in the ODE module is generated and governed by:
\begin{equation}
\frac{dh(t)}{dt} = f_\theta(h(t), t),
\end{equation}
where \( f_\theta \) is a neural network parameterized by \(\theta\) that defines the dynamics of the system. The hidden state \( h(t) \) at a future time \( t_1 \) is obtained by solving the initial value problem:
\begin{equation}
h(t_1) = h(t_0) + \int_{t_0}^{t_1} f_\theta(h(t), t) \, dt,
\end{equation}
which is approximated numerically using the ODE solver:
\begin{equation}
h(t_1) = \text{ODEInt}(f_\theta, h(t_0), [t_0, t_1]),
\end{equation}
where the function \(\text{ODEInt}\) computes the current hidden state, i.e., solution of the ODE over the time interval \([t_0, t_1]\).

\subsubsection{GRU-based Sequential Updates}
For each sequence, we initialize the hidden state as \( h_0 \). Then we iteratively update it using the observations \(\mathbf{x}_i\) at each time step with a gated recurrent unit (GRU)~\cite{zhu2024mci,cheng2018modeling}. The sequence update combines ODE dynamics with a GRU to handle the final observations, which is formulated as:
\begin{equation}
h(t_j) = \text{ODEInt}(f_\theta, h(t_{j-1}), [t_{j-1}, t_j]),
\end{equation}
\begin{equation}
h_j = \text{GRUCell}(\mathbf{x}_j, h(t_j)),
\end{equation}
where \( h(t_j) \) is the hidden state after solving the ODE, and \( h_j \) is the updated state incorporating the observation \(\mathbf{x}_j\) via the GRU cell.

\subsubsection{Mini-batch Encoding}
Given a batch of transaction sequences, the model processes each transaction independently. The final hidden states \( h_{i,T} \) for all transactions are aggregated into a batch representation, which is formulated as follows:
\begin{equation}
H_{\text{batch}} = \{h_{1,T}, h_{2,T}, \ldots, h_{N,T}\},
\end{equation}
where \( h_{i,T} \) represents the last hidden state of the \(i\)-th sequence, and \( N \) is the batch size. The last hidden state will be used as the input of the GNN-based classifier.
Overall, the neural ODE is responsible for modeling dynamics and generating hidden variable candidates. Then the gated recurrent unit cells are leveraged to generate neural representations incorporated with temporal information.

\subsection{Pseudo-Labeled Graph Generation}
To effectively utilize unlabeled data, we propose a dynamic labeling mechanism integrated with the Pre-Trained Beta Wavelet Graph Neural Network (BWGNN) \cite{Tang2022RethinkingGN}. Leveraging its robust classification capabilities, BWGNN serves as the backbone for generating and refining pseudo labels, ensuring an effective and adaptive labeling process that enhances the model's robustness and generalization.

The process begins with the transformation of the node features \( X \) through a Multi-Layer Perceptron (MLP), which captures the intrinsic node characteristics. The raw node features $X_{\text{raw}}$ were concatenated by the last hidden state $H_{\text{batch}}$ of the encoding module, which is formulated as: $X = X_{\text{raw}}||H_{\text{batch}}$. These features are then processed by a set of Beta wavelet kernels \( \mathcal{W}_{i, C-i} \), each designed to extract spectral information at specific frequency scales, resulting in transformed representations \( Z_i \):
\begin{equation}
Z_i = \mathcal{W}_{i, C-i}(\text{MLP}(X)), \quad i \in \{0, 1, \ldots, C\}.
\end{equation}

The outputs from the wavelet kernels are aggregated into a comprehensive node representation \( H \):
\begin{equation}
H = \text{AGG}([Z_0, Z_1, \ldots, Z_C]),
\end{equation}
where \( \text{AGG} \) combines multi-scale spectral features. This representation is further processed by another MLP with a Sigmoid activation to compute anomaly probabilities \( p_i \) for each node:
\begin{equation}
p_i = \text{Sigmoid}(\text{MLP}(H)).
\end{equation}

Based on these probabilities, pseudo labels are generated by applying a threshold \( z \), which is formulated as follows:
\begin{equation}
\hat{y}_i = 
\begin{cases} 
1 & \text{if } p_i > z \\
0 & \text{otherwise}.
\end{cases}
\end{equation}

The transaction labeling mechanism provides a fully labeled graph for each batch of nodes to have graph data for heterogeneous graph attention during the training process.

\subsection{Heterogeneous Graph Attention}

In multi-relational graphs, redundant features can hinder the learning process by introducing noise. To address this challenge, we adopt a heterogeneous graph attention mechanism that consists of two main components: \textbf{intra-attention} and \textbf{inter-attention}. This approach efficiently handles the complexity of multi-relational transaction graphs by attending to information both within individual nodes and between different types of nodes, ensuring all meaningful relationships are captured.

\subsubsection{Intra-Attention}

The intra-attention mechanism is designed to consolidate features from neighboring nodes connected by the same relation type. This ensures that the model can choose important neighbors and capture shared characteristics within each type of node. The process includes the following steps:

1. \textit{Attention-based Neighbor Aggregation} within each relation:
\begin{equation}
h_{v, r,*}^{\prime, l} = \sum_{u \in \mathcal{N}_{r,*}(v)} \alpha_{vu, r} h_u^{l-1}
\end{equation}

The attention weights \( \alpha_{vu, r} \) are calculated using the softmax function to normalize the importance of neighboring nodes:
\begin{equation}
\alpha_{vu, r} = \frac{\exp\left(\text{LeakyReLU}\left(a_r^\top \cdot \left[W_r^l h_v^{l-1} \parallel W_r^l h_u^{l-1}\right]\right)\right)}{\sum_{k \in \mathcal{N}_{r,*}(v)} \exp\left(\text{LeakyReLU}\left(a_r^\top \cdot \left[W_r^l h_v^{l-1} \parallel W_r^l h_k^{l-1}\right]\right)\right)}
\end{equation}

Here, \( a_r \) is a learnable vector specific to the relation \( r \), \( W_r^l \) is a relation-specific weight matrix, and \( \parallel \) denotes concatenation.

2. \textit{Feature Update} using attended neighbor features:

\begin{equation}
h_{v, r,*}^l = \text{ReLU}\left(W_{\text{intra}, r}^l \cdot \left(h_v^{l-1} \parallel h_{v, r,*}^{\prime, l}\right)\right)
\end{equation}
where \( W_{\text{intra}, r}^l \) is the learnable weight matrix for intra-attention at layer \( l \), \( h_v^{l-1} \) represents the central node's features at the previous layer, \( h_{v, r,*}^{\prime, l} \) is the aggregated feature from the neighbors under relation \( r \), and \( \text{ReLU} \) introduces non-linearity to the updated features.

The intra-attention mechanism focuses on leveraging homogeneous node relationships among the same type of nodes to refine the neighbor node feature representations.

\subsubsection{Inter-Attention}

After introducing intra-attention, the model proceeds to inter-attention, where it collects information across different types of nodes. The process is as follows: (1) Features within each type are aggregated using a mean operation, ensuring each type’s features are combined based on their relationships; (2) The attended features from different types of nodes are then passed through an attention mechanism to compute the attention weights, allowing the model to focus on the most informative relationships between the central node and different types of relationships.

The inter-attention process is performed as follows:

1. \textit{Mean Aggregation} within each relation:
\begin{equation}
h_{v, r,g}^{\prime, l} = \text{Agg}_{\text{mean}}(h_u^{l-1}), \quad \forall u \in \mathcal{N}_{r,g}(v)
\end{equation}
where \( \mathcal{N}_{r,g}(v) \) represents the set of neighboring nodes of \( v \) connected by relation \( r \) and group \( g \), \( \text{Agg}_{\text{mean}} \) denotes the function to compute the mean values of neighboring node features \( h_u^{l-1} \), and \( h_u^{l-1} \) is the feature of neighbor \( u \) at layer \( l-1 \).

2. \textit{Inter-Attention} between different relations:
\begin{equation}
h_v^l = \sum_r \sum_g \alpha_{r,g}^l h_{v,r,g}^l
\end{equation}
where \( h_v^l \) is the updated feature of the central node \( v \) at layer \( l \), \( h_{v,r,g}^l \) is the feature aggregated from relation \( r \) and group \( g \), and \( \alpha_{r,g}^l \) is the attention weight for relation \( r \) and group \( g \).
The attention weights \( \alpha_{r,g}^l \) are computed using the softmax function:
\begin{equation}
\alpha_{r,g}^l = \frac{\exp(\omega_{r,g}^l)}{\sum_m \exp(\omega_{r,g}^l)}
\end{equation}

Here, \( \omega_{r,g}^l \) is the unnormalized attention score for relation \( r \) and group \( g \), and \( \sum_m \) normalizes the scores across all relations and groups.
The weight \( \omega_{r, g}^l \) is determined by the interaction between the node's features and the attended features from each relation:
\begin{equation}
\omega_{r, g}^l = q^T \cdot \tanh(W_{\text{inter 1}}^l h_v^{l-1} + W_{\text{inter 2}}^l h_{v, r, g}^l)
\end{equation}

In this equation, \( q \) is a learnable parameter vector that projects the interaction into a scalar, \( W_{\text{inter 1}}^l \) and \( W_{\text{inter 2}}^l \) are learnable weight matrices for the central node and the relation-specific features, respectively, and \( \tanh \) introduces non-linearity to the interaction.
The inter-attention mechanism ensures that information is aggregated across relations and groups, enabling the model to capture complex dependencies between different types of nodes and relations.
After both intra-attention and inter-attention, the final node representation is obtained by concatenating the features from all relational representations with the central node's feature. This step ensures that the model incorporates information from both the central node and the different relationships:
\begin{equation}
h_v^{\text{final}} = h_v^{l} \parallel \text{concat} \left( \{h_{v, r,g}^l\}_{r,g} \right)
\end{equation}

\subsection{Optimization Objective}

After the graph attention process, the embedding \( h_v^{\text{final}} \) obtained from the final layer is input into a Multilayer Perceptron (MLP), which produces a classification score \( p_v \) representing the probability of node \( v \) belonging to each category. The probabilities are computed using the softmax function, enabling the model to estimate the likelihood for each class. The training process minimizes the cross-entropy loss, which is defined as:
\begin{equation}
\mathcal{L} = -\sum_{v \in \mathcal{V}} \sum_{c=1}^{C} y_{v,c} \log p_{v,c},
\end{equation}
where \( y_{v,c} \) represents the ground-truth label for node \( v \) in class \( c \), \( p_{v,c} \) is the predicted probability, and \( C \) is the total number of classes.

At the end of each training epoch, for spoofing detection, i.e., binary classification tasks, a threshold \( z \) is often used for decision-making, where the prediction is determined as follows:
\begin{equation}
\hat{y}_v = 
\begin{cases} 
1 & \text{if } p_v > z, \\
0 & \text{otherwise.}
\end{cases}
\end{equation}

This binary decision and prediction probability are used for model performance comparison experiments and building real-world spoofing detection systems.

\section{Experiments}
\label{sec:exp}
In this section, we first show the detail of experimental settings. Then we report the performance comparison results on spoofing detection task. After that, we introduce the ablation study, parameter sensitivity, case studies, and implementations.

\subsection{Experimental Settings}
\subsubsection{Datasets}
We have created a new dataset called Spoofing Detection Dataset, consisting of 40,072 transaction records collected from our partners between January 5, 2018, and November 14, 2018. The dataset contains 49 feature dimensions, which can be categorized into four types of information: order-related details (e.g., order price, order balance, and order date), market and price data (e.g., today's trading volume and value), as well as order positions and profit or loss figures. The ground truth labels are based on cases reported by traders and verified by financial domain experts. Transactions are labeled as 1 if identified as fraudulent, and 0 otherwise.

During the data preprocessing stage, we remove irrelevant columns, such as trader ID and customer ID. Then, the features are normalized based on the train set statistics. To construct the graph, each transaction is treated as a node. Edges between nodes are established using a sliding window based on the transaction date.

\subsubsection{Compared Methods}
The compared methods can be classified into two parts: (1) Traditional learning-based methods; and (2) Advanced Spoofing Detection Methods. 
The traditional learning-based methods include Logistic Regression (LR) \cite{LR}, Random Forest (RF) \cite{RF}, Adaboost \cite{Adaboost}, Gradient Boosting Decision Tree (GBDT) \cite{GBDT}, Hybrid Multi-layer Perceptron (HMLP) \cite{HMLP}, Long Short Term Memory (LSTM) \cite{LSTM}, and BiTransformer \cite{BiTransformer}, all of which are implemented with their default hyperparameter settings.

For the advanced spoofing detection methods, we compare with EigenGCN \cite{EigenGCN}, a graph convolutional network designed for transaction relationship learning; RetaGNN \cite{RetaGNN}, a relational temporal attention-based graph neural network; GRU-DM \cite{GRU-DM}, a Gated Recurrent Unit framework for spoofing detection using market indicators; RTG-Trans \cite{RTG-Trans}, a temporal gating method for detecting dynamic interactions within spoofing detection graphs; GPEGNN \cite{GPEGNN}, a multi-layer graph attention-based method for local and global context learning; and GDGM, our proposed method, with hyperparameter settings detailed in section~\ref{subsec:param}.

\subsubsection{Evaluation Metrics}
To evaluate the the performance of our model on spoofing detection, we leverage five widely recognized metrics to assess: Area Under the ROC Curve (AUC), Precision, Recall, F1 Score and Accuracy.
The AUC measures evaluates the model's ability to differentiate between classes at various threshold settings. 
Precision ($P$) measures the proportion of true positive predictions among all positive predictions. It is computed as:
$P = \frac{N_{TP}}{N_{TP} + N_{FP}}$
where $N_{TP}$ represents the number of true positives, and $N_{FP}$ represents the number of false positives.
Recall ($R$) quantifies the model's ability to identify all true positive instances. It is calculated as: $R = \frac{N_{TP}}{N_{TP} + N_{FN}}$
where $N_{FN}$ denotes the number of false negatives.
The F1 Score provides a harmonic mean of Precision and Recall, balancing the trade-off between these two metrics. It is expressed as: $F1 = \frac{2 \times P \times R}{P + R}$
Accuracy evaluates the overall correctness of the model by considering both positive and negative classes. It is defined as: $Accuracy = \frac{N_{TP} + N_{TN}}{N_{TP} + N_{TN} + N_{FP} + N_{FN}}$ where $N_{TN}$ denotes the number of true negatives.


\begin{figure*}[t]
  \centering
  \includegraphics[width=\linewidth]{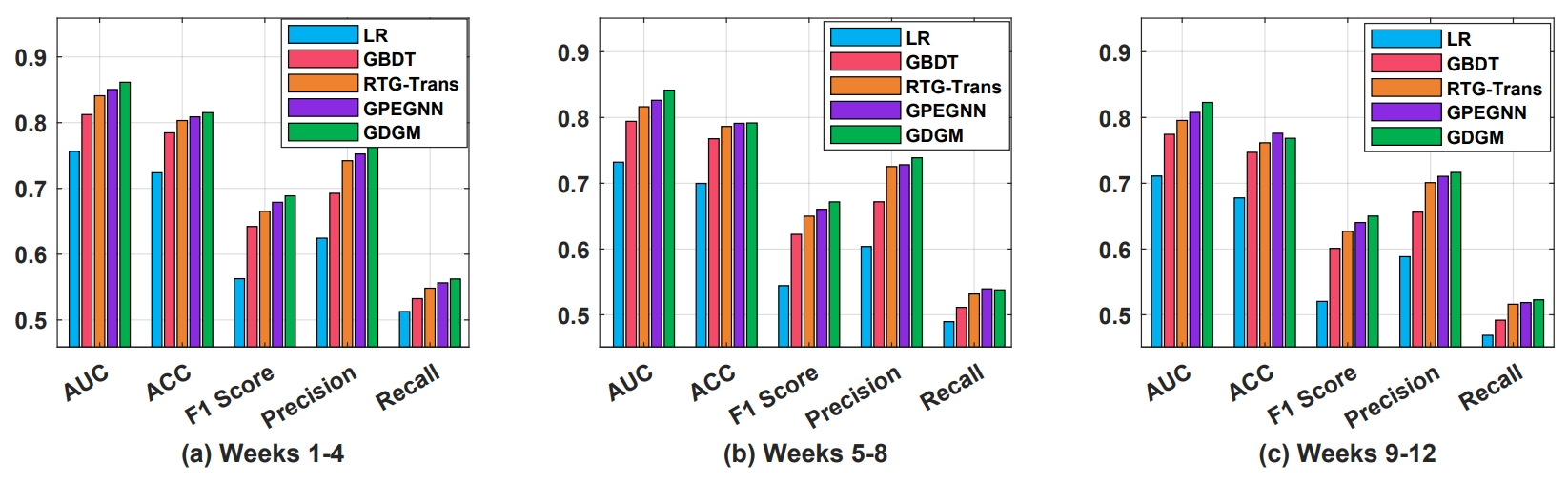}
  \caption{The experimental results of spoofing detection methods under queue-based study. During the experiment, pre-trained models were employed to detect spoofing transactions over four weeks. At the end of each four-week interval, newly observed cases were merged with the historically labeled database, and the models were retrained to improve their performance.}\label{fig:imple_exp}
\end{figure*}

\subsection{Performance Comparison}
In this section, we evaluate the performance of our proposed model against various baseline methods on the spoofing detection task. Table~\ref{tab:performance_metrics} provides a comprehensive comparison using metrics such as Area Under the ROC Curve (AUC), Accuracy, F1 Score, Precision, and Recall. Each model was tested over multiple iterations, and the average results are reported.
The first six rows of Table~\ref{tab:performance_metrics} present traditional machine learning methods, including Logistic Regression (LR), Random Forest (RF), and gradient boosting models (Adaboost and GBDT). Among these, RF achieves the highest AUC (0.8985), significantly outperforming other traditional methods. However, these baseline models, while effective, demonstrate limitations in capturing complex relational patterns, leading to suboptimal recall values, particularly for LR (0.4262) and Adaboost (0.5568).
The middle section of Table~\ref{tab:performance_metrics} introduces more advanced graph-based and sequence-based models, such as BiTransformer, EigenGCN, and RTG-Trans. These methods show marked improvements over traditional baselines, with BiTransformer achieving an AUC of 0.8957 and EigenGCN delivering competitive precision (0.8059). Notably, RTG-Trans outperforms most other baselines in recall (0.7507), indicating its effectiveness in detecting spoofing cases with minimal false negatives. However, even these models exhibit limitations in balancing all performance metrics, as seen in slightly lower F1 scores for some methods.
Our proposed model, listed as ``Ours'' in Table~\ref{tab:performance_metrics}, achieves the best overall performance across all metrics. It records the highest AUC (0.9029), Accuracy (0.8701), and F1 Score (0.8002), along with superior Precision (0.8508) and Recall (0.7702). These results demonstrate the effectiveness of our approach in leveraging intricate graph-based relationships and temporal dependencies for spoofing detection. Compared to the best-performing baseline (RTG-Trans), our model achieves a 0.31\% improvement in AUC and a 2.4\% increase in Accuracy, showcasing its robustness and reliability in handling complex spoofing scenarios.
This significant performance boost highlights the capability of our model to capture nuanced patterns and relationships in graph-structured data, making it particularly well-suited for detecting spoofing and other fraudulent activities in real-world applications.

\begin{table}[t]
    \centering
    \caption{Experimental results for different machine learning methods on spoofing detection task.}
    \resizebox{0.5\textwidth}{!}{%
    \begin{tabular}{l|c|c|c|c|c}
        \hline
        \textbf{Method} & \textbf{AUC} & \textbf{Accuracy} & \textbf{F1 Score} & \textbf{Precision} & \textbf{Recall} \\
        \hline
        LR & 0.7828 & 0.8119 & 0.5320 & 0.7077 & 0.4262 \\
        RF & 0.8985 & 0.8578 & 0.7066 & 0.7322 & 0.6826 \\
        Adaboost & 0.8716 & 0.8538 & 0.6564 & 0.7994 & 0.5568 \\
        GBDT & 0.8911 & 0.8615 & 0.6720 & 0.8284 & 0.5653 \\
        HMLP & 0.7587 & 0.7893 & 0.5894 & 0.7785 & 0.5852 \\
        LSTM & 0.7759 & 0.8393 & 0.7483 & 0.8149 & 0.7058 \\
        \hline
        BiTransformer & 0.8957 & 0.8529 & 0.7504 & 0.8109 & 0.7205 \\
        EigenGCN & 0.8904 & 0.8491 & 0.7405 & 0.8059 & 0.7102 \\
        RetaGNN & 0.8935 & 0.8552 & 0.7705 & 0.8201 & 0.7409 \\
        GRU-DM & 0.8805 & 0.8483 & 0.7601 & 0.8152 & 0.7308 \\
        RTG-Trans & 0.8998 & 0.8602 & 0.7807 & 0.8255 & 0.7507 \\
        GPEGNN & 0.8989 & 0.8578 & 0.7852 & 0.8309 & 0.7558 \\
        \hline
        Ours & \textbf{0.9029} & \textbf{0.8701} & \textbf{0.8002} & \textbf{0.8508} & \textbf{0.7702} \\
        \hline
    \end{tabular}
    }
    \label{tab:performance_metrics}
\end{table}

\subsection{Implementation and Online Deployment}
In addition to the comparison on historical data, testing the accuracy on future transactions is more important to real-world applications.
As to real-world deployment of spoofing detection, transactions are processed through a distributed message queue for real-time evaluation. Initially, they are checked against blacklist entries and fraud detection rules (in-process detection). Transactions that hit any blacklist or fraud rule are blocked immediately. If no matches are found, user and symbol features are then extracted and passed to an online predictive model (GDGM, in this case) as part of the post-process detection.
Throughout this process, historical transaction data flagged by the detection systems is stored in an in-memory database to support large-scale detection. High-risk transactions are escalated to domain experts for verification, with feedback on these cases stored in the historical database. The predictive model is periodically retrained in batches based on the latest expert feedback, allowing it to learn from recent spoofing patterns. Newly identified fraud cases are also incorporated to refine the blacklist and fraud detection rules, ensuring both detection layers remain adaptive and robust.
To evaluate the performance of our model under real-world conditions, we tested 105 confirmed spoofing cases using data collected between July and September through a 12-weeks queue-based study. We compared our proposed GDGM model with two widely-used baselines, Logistic Regression (LR) and Gradient Boosting Decision Tree (GBDT), as well as two state-of-the-art (SOTA) methods specifically designed for spoofing detection, RTG-Trans and GPEGNN. The comparison results of the online experiments are summarized in Figure~\ref{fig:imple_exp}.
As shown in Figure~\ref{fig:imple_exp}, in the first 4 weeks, GDGM consistently outperforms all baseline and state-of-the-art methods across all evaluation metrics. Notably, GDGM achieves the highest AUC (0.8614), ACC (0.8152), F1 score (0.6887), precision (0.7624), and recall (0.5623). These results highlight the ability of our model to effectively identify spoofing cases in real-world settings, even when dealing with noisy or irregular data. Similar conclusions can be derived from the experimental results of the next 8 weeks.
The superior performance of GDGM can be attributed to its ability to model irregular trading behaviors and heterogeneous relationships among transactions, as well as its integration of pseudo-labeling and graph aggregation techniques. By leveraging these advanced capabilities, GDGM captures subtle patterns indicative of conspiracy spoofing that are often missed by traditional baselines and even existing SOTA methods.
Furthermore, the deployment of GDGM in a real-time production environment demonstrates its practicality and scalability. The integration with a distributed message queue ensures low-latency processing, while the adaptive learning framework enables the model to evolve based on the latest spoofing trends. These features make GDGM a robust and reliable solution for combating financial fraud in dynamic, high-stakes trading markets.

\begin{figure}[tb!]
\includegraphics[width=3.4in]{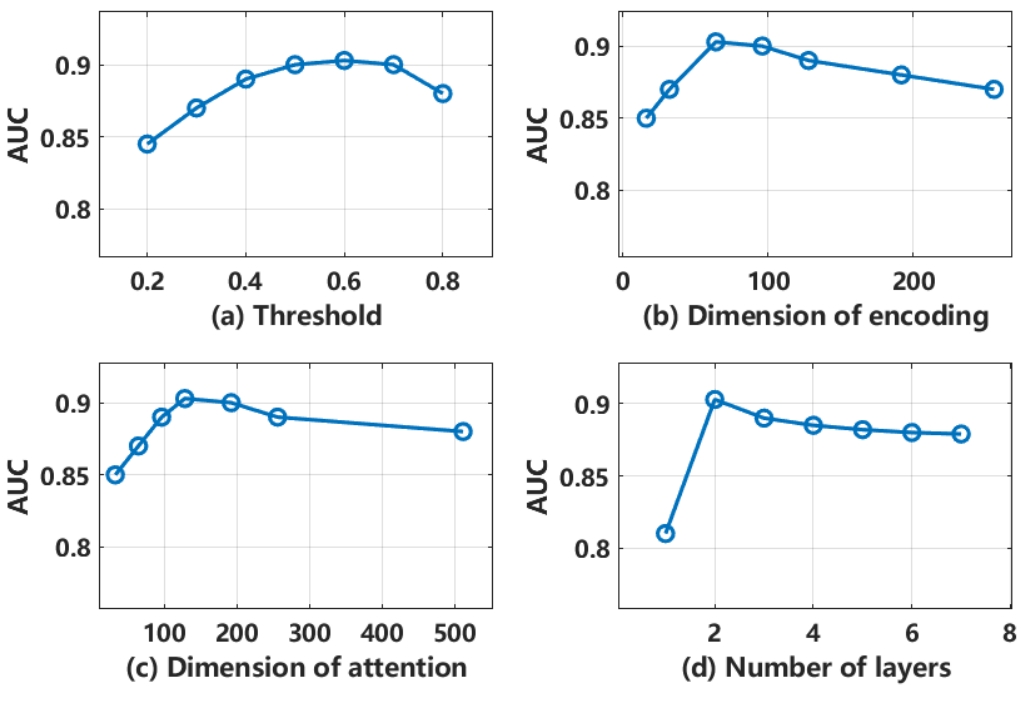}
\caption{
AUC of our method, in terms of threshold $z$, dimension of encoding output $h$, dimension of the attention vector $q$, and number of heterogeneous aggregation layers.
}
\label{fig:param}
\centering
\end{figure}

\subsection{Parameter Sensitivity}
\label{subsec:param}
One of the important research questions is to analyze the sensitivity of our model to its hyperparameters. In this experiment, we examine key parameters such as the threshold for pseudo-labeling, the dimension of encoding output, the dimension of the attention vector, and the number of graph aggregation layers. 
Specifically, Figure \ref{fig:param}(a) shows the effect of varying the threshold for pseudo-labeling on model performance. As the threshold increases from 0.2 to 0.8, the AUC metric steadily improves, peaking when the threshold is set to 0.6. Figure \ref{fig:param}(b) evaluates the sensitivity to the dimension of encoding output. Encoding output dimension shows best performance at 64 dimensions, with higher values leading to degradation likely from overfitting. The impact of the dimension of the attention vector is shown in Figure \ref{fig:param}(c). The AUC steadily improves as the dimension increases from 32 to 128, reaching a peak. Beyond this point, larger attention vectors result in marginally lower performance, indicating that 128 is an optimal choice. Figure \ref{fig:param}(d) examines the effect of the number of graph aggregation layers. The performance improves as the number of layers increases from 1 to 2, achieving the highest AUC at 2 layers. More layers lead to a decline in performance, suggesting that excessive stacking of graph aggregation layers may introduce noise or redundancy.
Based on these findings, we select the best parameter settings as follows: a threshold of 0.6 for pseudo-labeling, an encoding output dimension of 64, an attention vector dimension of 128, and 2 graph aggregation layers. These configurations enable our model to achieve optimal performance on the spoofing detection task.

\begin{table}[t]
    \centering
    \caption{Ablation study results, showing the impact of different model components on spoofing detection task.}
    \begin{tabular}{l|c|c|c}
        \hline
        \textbf{Model Variant} & \textbf{AUC} & \textbf{F1 Score} & \textbf{Precision} \\
        \hline
        GDGM & \textbf{0.9029} & \textbf{0.8002} & \textbf{0.8508} \\\hline
        w/o Pseudo Label & 0.8998 & 0.7807 & 0.8255 \\
        w/o Neural ODE & 0.8965 & 0.7513 & 0.8126 \\
        w/o Hete. GNN & 0.8901 & 0.7413 & 0.8041 \\
        \hline
    \end{tabular}
    \label{tab:ablation_study}
\end{table}

\subsection{Ablation Study}

To evaluate the contribution of each component in our model, we conducted an ablation study by removing specific features from the model: \textbf{(1) Without Pseudo Label:} This variant directly uses the training set labels as input to the heterogeneous graph neural network instead of pseudo labels, which reduces the model's ability to leverage inferred label consistency. \textbf{(2) Without Neural ODE:} This configuration replaces our ODE-RNN encoder with a standard RNN encoder, limiting the model's capacity to capture continuous temporal dynamics effectively. \textbf{(3) Without Heterogeneous GNN:} In this variant of the model, instead of employing the heterogeneous graph neural network, we chose to use a homogeneous graph neural network, specifically the Graph Attention Network (GAT), for the classification task. When we switch to a homogeneous GNN like GAT, the model's expressiveness in accurately representing and modeling these diverse node and edge types is reduced. 

The experimental results obtained from the spoofing detection dataset, which are clearly presented in Table \ref{tab:ablation_study}, demonstrate the significance of each of these components. The full model, which incorporates all the components including the pseudo labels, neural ODEs, and heterogeneous GNNs, achieves the best performance across all the metrics that were considered for evaluation. This outstanding performance of the full model serves to highlight the importance of integrating these specific components together. Notably, when we replace the pseudo label mechanism, we observe a moderate decline in the performance of the model. Moreover, when either the neural ODE encoder or the heterogeneous GNN is removed from the model, the effectiveness of the model is further degraded to an even greater extent. These experimental outcomes clearly underline the critical role that each component plays in enabling the model to achieve robust spoofing detection capabilities.

\begin{figure}[tb!]
\includegraphics[width=3.4in]{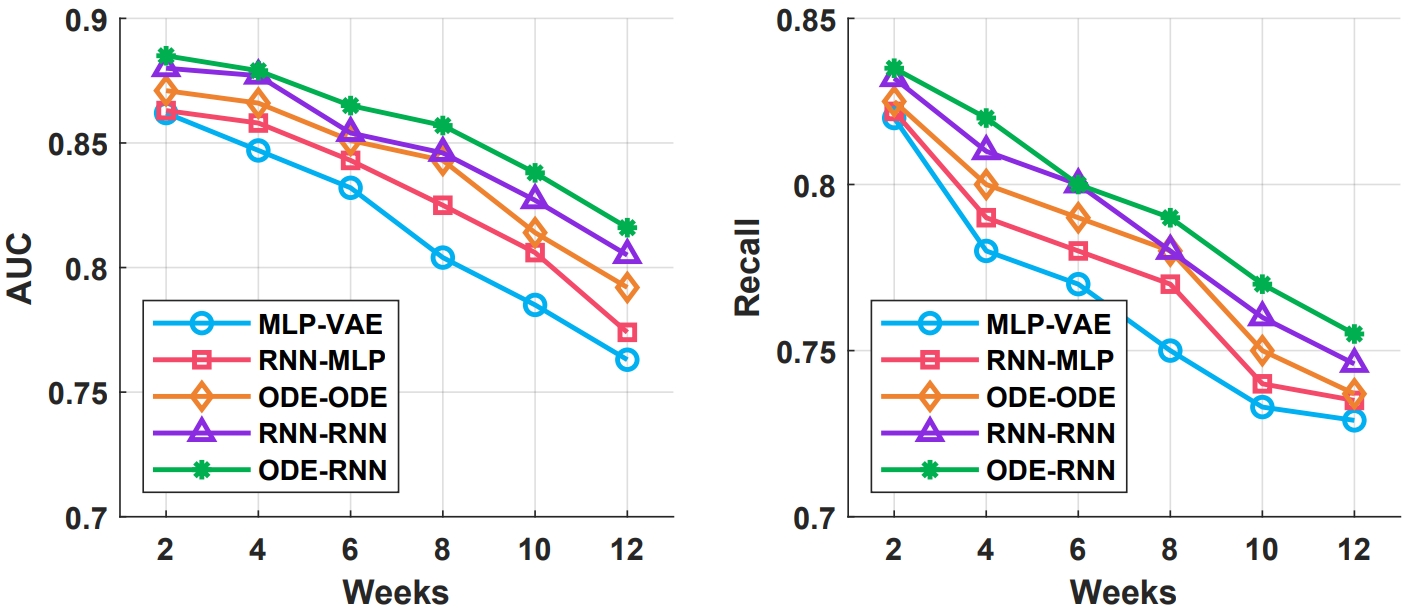}
\caption{
Performance comparison for models with different generative data encoding modules. We fix the trained model and make predictions for the next 12 weeks.
}
\label{fig:encoding_comparison}
\centering
\end{figure}

In addition to the ablation components that were mentioned above, we also extended our investigation to explore the performance of various encoding modules that are commonly used in the context of spoofing detection. The experimental comparison of these different encoding modules is visually depicted in Figure \ref{fig:encoding_comparison}. As can be clearly observed from the figure, the ODE-RNN encoding module outperforms all the other models that were included in the comparison. Following closely behind in terms of performance is the RNN-RNN model, and then the Neural ODE. On the other hand, the MLP-VAE and RNN-MLP models display relatively weaker performance. This is primarily due to their inherent inability to fully capture both the temporal and sequential dynamics of the data. These dynamics are essential for accurately understanding and processing the information within the spoofing detection dataset, and the failure to effectively capture them results in the observed suboptimal performance of these particular encoding modules.


\section{Conclusion}
Spoofing detection in financial trading, particularly intricate behaviors like conspiracy spoofing, is a critical yet challenging task. Conventional machine learning methods often neglect the interconnected and heterogeneous nature of trading data, while existing graph-based techniques struggle to capture the irregularities inherent in real-world trading behaviors. Therefore, we proposed the Generative Dynamic Graph Model (GDGM), a novel framework designed to model both temporal trading behaviors and dynamic, heterogeneous relationships among nodes.
Our approach leverages neural ordinary differential equations and gated recurrent units to represent irregular trading patterns. Then, we employ the pre-trained model for pseudo-labeling and heterogeneous attention-based aggregation mechanisms to effectively capture conspiratorial spoofing signals. The results of extensive experiments demonstrate that GDGM outperforms state-of-the-art models in detecting spoofing behaviors, highlighting its effectiveness and robustness. Furthermore, the successful deployment of our system in one of the largest global trading markets underscores its practical applicability, validating its exceptional performance in real-world scenarios.

\begin{acks}
The work is supported by the National Natural Science Foundation of China (62472317), the Fundamental Research Funds for the Central Universities and the Shanghai Science and Technology Innovation Action Plan Project (Grant no. 22YS1400600 and 24692118300).
\end{acks}

\bibliographystyle{www25}
\bibliography{www25}


\end{document}